\documentclass[11pt,twoside]{article}
  \pdfoutput=1

\usepackage{amsmath}
\usepackage{amssymb}
\usepackage{fancyhdr}
\usepackage{graphicx}
\usepackage{subfigure}

\usepackage[textwidth=16cm,textheight=22cm,top=3cm,bottom=3cm]{geometry}

\graphicspath{{./figuras_nuevas2/}}

 \newtheorem{theorem}{Theorem}
 \newtheorem{remark}{Remark}
 \newtheorem{corollary}{Corollary}

\newcommand{\drop}[1]{}
\newcommand{\no}{\noindent}
\newcommand{\fer}[1]{(\ref{#1})}
\newcommand{\qtext}[1]{\quad\text{#1}}

\newcommand{\bx}{\mathbf{x}}
\newcommand{\by}{\mathbf{y}}

\newcommand{\cK}{\mathcal{K}}

\newcommand{\N}{\mathbb{N}}
\newcommand{\R}{\mathbb{R}}

\def\O{\Omega}

\newcommand{\abs}[1]{| #1 |}
\newcommand{\nor}[1]{\| #1 \|}

\DeclareMathOperator{\NF}{NF}
\DeclareMathOperator{\BF}{BF}
\DeclareMathOperator{\NL}{NL}

\title{Neighborhood filters and the decreasing rearrangement
\thanks{Supported by Spanish MCI Project MTM2010-18427.}}

\author{Gonzalo Galiano  \thanks{Dpt. of Mathematics, Universidad de Oviedo,
 c/ Calvo Sotelo, 33007-Oviedo, Spain ({\tt galiano@uniovi.es, julian@uniovi.es})}
    \and Juli\'an Velasco\footnotemark[2] }
\date{}

\begin{document}

\date{Final version: J. Math Imaging Vision\\
DOI 10.1007/s10851-014-0522-3}

\maketitle

\begin{abstract}
Nonlocal filters are simple and powerful techniques for image denoising.
In this paper, we give new insights into the analysis of one kind of them, 
the Neighborhood filter, by using a classical although not commonly used
transformation: the decreasing rearrangement of a function.
Independently of the dimension of the \emph{image}, we reformulate 
the Neighborhood filter and its iterative variants as 
an integral operator defined in a one-dimensional space. The simplicity of
this formulation allows to perform a detailed analysis of its properties.
Among others, we prove that the filtered image is a contrast change of the original image, an that the filtering procedure behaves asymptotically as a 
shock filter combined with a border diffusive term, responsible for 
the staircaising effect and the loss of contrast.

\no\textbf{keywords: }Neighborhood filters, decreasing rearrangement, denoising, segmentation
\end{abstract}


\section{Introduction}

Let $\O\subset\R^d$ $(d\geq 1)$ be an open and bounded set,
and consider a function $u\in L^\infty(\O)$. 
The Neighborhood filter operator  is defined by
\begin{equation}
\label{def.NF}
 \NF^h u (\bx)=\frac{1}{C(\bx)}\int_\O \textrm{e}^{-\frac{\abs{u(\bx)-u(\by)}^2 }{h^2}} u(\by)d\by,
\end{equation}
where $h$ is a positive constant, and \[C(\bx)=\int_\O \exp\left(-\abs{u(\bx)-u(\by)}^2) h^{-2}\right) d\by\] is a normalization factor, intended to allow constants to be fixed points of $\NF^h$. 

The Neighborhood filter (NF) is the simplest particular case of other related filters involving local terms, 
notably the Yaroslavsky filter \cite{Yaroslavsky1985,Yaroslavsky2003}, the SUSAN filter \cite{Smith1997} introduced by Smith and Brady, the Bilateral filter \cite{Tomasi1998} of Tomasi and Manduchi, and the Nonlocal Means filter (NLM) \cite{Buades2005} by Buades, Coll and Morel.

These methods have been introduced in the last decades
as efficient alternatives to local methods such as those expressed in terms 
of nonlinear diffusion partial differential equations (PDE's), among which the pioneering approaches of Perona and Malik \cite{Perona1990}, \'Alvarez, Lions and Morel \cite{Alvarez1992} and Rudin, Osher and Fatemi \cite{Rudin1992} are fundamental. 
We refer the reader to \cite{Buades2010} for a review and comparison of these methods.

Among all these filters, the Neighborhood filter is the simplest, but yet useful, 
method due to its compromise between denoising quality and computational speed. 
Indeed, although it creates shocks and staircasing effects \cite{Buades2006a}, the computational cost 
is by far lower than those of other integral kernel filters or PDE's based methods.

Since, usually, a single denoising step of the nonlocal filters is not enough, 
an iteration  is performed according to several choices of 
the iteration actualization, see \fer{def.GFF} and \fer{def.GFV} for two of such 
strategies.

In this context, the Neighborhood filter and its variants have been analyzed
from different points of view. For instance, Barash \cite{Barash2002}, Elad \cite{Elad2002},  Barash et al. \cite{Barash2004}, and Buades et al. \cite{Buades2006}
investigate the asymptotic relationship between the Yaroslavsky filter and the Perona-Malik PDE. Gilboa et al. \cite{Gilboa2008} study certain applications of 
nonlocal operators to image processing. In \cite{Peyre2008}, Peyr\'e establishes a relationship
between the non-iterative nonlocal filtering schemes and thresholding in adapted
orthogonal basis.
In a more recent paper, Singer et al. \cite{Singer2009}
interpret the Neighborhood filter as a stochastic diffusion process,
explaining in this way the attenuation of high frequencies in the processed images.

In this article, we reformulate the Neighborhood filter in terms of the \emph{decreasing rearrangement} of the initial image, $u$,  
which is defined as the inverse of the \emph{distribution function}
$q\in\R\to m_u(q) = \abs{\{\bx \in\O : u(\bx) >q\}}$,
see Section 2 for the precise definition.

Realizing that the structure of level sets of $u$ is invariant through the Neighborhood filter operation as well as through the decreasing rearrangement of 
$u$ allows us to rewrite \fer{def.NF} in terms of a one-dimensional integral 
expression, see Theorem~\ref{th.equivalence1}.

Although from expression \fer{def.NF} is readily seen that only computation on level lines is needed to perform the filtering, the alternative expression in terms of the decreasing rearrangement offers room for further analysis of the iterative scheme.

Perhaps, the most important consequence of the rearrangement is, apart from 
the dimensional reduction, the reinterpretation of the NF as a 
\emph{local} algorithm. Thanks to this, we may prove  some properties of the Neighborhood filter nonlinear iterative scheme, see \fer{def.GFV}, among which
\begin{itemize}
 \item The asymptotic behavior of the NF as a shock filter of the type introduced by \'Alvarez et al. \cite{Alvarez1994}, combined with a contrast loss effect, see Theorem~\ref{th.pde}.
 
 \item The contrast change character of the NF, i.e. the existence of a continuous and
 increasing function $g:\R\to\R$ such that $\NF^h (u(\bx))=g(u(\bx))$, see Corollary~\ref{cor.contrast}.
\end{itemize}

As mentioned above, the most salient advantage of the Neighborhood filter in its rearranged version is speed. If $N$ denotes the image size in pixels, the complexity of other
nonlocal filters such as the \emph{classical} NF, the Bilateral or the NLM filters are of the order $C\times N$, 
where $C$ depends on different parameters (window sizes) used in those models. A typical value of $C$ may be of the order $10^5$, for the NLM. However, the complexity of the rearranged version of the NF depends on the number of the initial image intensity levels, which we assume quantized in $Q$ levels, and on some small constant related to the window size, $h$, resulting 
in a complexity of the order $c\times Q$, being a typical value of $c$ of the order $10^2$.
Thus, the complexity of the NF is independent of the image size, once the level lines of the
initial image have been identified.

However, denoising quality of the NF is, by far, poorer. The staircaising 
effect, always present in algorithms reducing the Total Variation of the initial image, is especially strong for the NF. 
In fact, after several iterations, and depending on the window size $h$, the NF output image concentrates most of its pixels mass on few level sets, producing a segmentation-like effect on the initial image.

A quick explanation of the difference between the NF and the Bilateral and NLM filters is that the NF does not retain any local information of the image, diffusing the
intensity values just according to the mass of their corresponding level lines. Thus, a pixel belonging to a level line with large mass will retain its value even if it is isolated in a component of the image with different intensity value. 

Due to this, the Neighborhood filter and morphological filters deduced from the topographic map of an image, see for instance the monograph by Caselles and Monasse \cite{Caselles2010}, such as the Grain and the Killer filters,  are also different since the latter use the local geometry of level sets connected components in a fundamental manner. 

However, they do share some similitudes in their level set based framework, and this could be used for combining both algorithms. For instance, the isolated small regions belonging to a level line with large mass which remain in the filtered image after 
the NF application could be removed by morphological procedures, such as 
opening and closing operators.

The segmentation-like behavior of the NF may be explained in terms of the relationship between
the inflexion points of the decreasing rearrangement and the local extreme points of the image histogram, $h(q)=-m'_u(q)$. Thus, since the NF behaves asymptotically as a shock filter, and shock filters accumulate mass in inflexion points, we deduce that the NF works as an histogram based segmentation algorithm. 

Since for our analysis the Gaussian form of the integral kernel is not important, we consider the following generalization of the Neighborhood filter
\begin{equation}
\label{def.GF}
 \NF^h u (\bx)=\frac{1}{C(\bx)}\int_\O \cK_h(u(\bx)-u(\by)) u(\by)d\by,
\end{equation}
with $C(\bx)=\int_\O \cK_h(u(\bx)-u(\by)) d\by$, and $\cK_h(\xi)= \cK(\xi/h)$. For the moment, we only assume 
$\cK \in L^1_{loc}(\R)$ and $\cK\geq0$ to have \fer{def.GF} well defined, although more meaningful conditions will be
stated later. 

We consider the following iterative schemes, 
for $n\in\N$ (including $n=0$), and $u^{(0)}=u$: 
\begin{enumerate}
 \item Iteration with fixed kernel (linear operator),
\begin{equation}
\label{def.GFF}
 u^{(n+1)} (\bx)=\frac{1}{C_0(\bx)}\int_\O \cK_h(u^{(0)}(\bx)-u^{(0)}(\by)) u^{(n)}(\by)d\by,
\end{equation}
 with $C_0(\bx)=\int_\O \cK_h(u^{(0)}(\bx)-u^{(0)}(\by)) d\by$.
 
 \item Iteration with varying kernel (nonlinear operator),
\begin{equation}
\label{def.GFV}
 u^{(n+1)} (\bx)=\frac{1}{C_n(\bx)}\int_\O \cK_h(u^{(n)}(\bx)-u^{(n)}(\by)) u^{(n)}(\by)d\by,
\end{equation}
 with $C_n(\bx)=\int_\O \cK_h(u^{(n)}(\bx)-u^{(n)}(\by)) d\by$.
\end{enumerate}
Observe that, for both schemes, we have
\begin{equation}
\label{bound.iter}
 \nor{u^{(n+1)}}_{L^\infty(\O)}\leq \nor{u^{(n)}}_{L^\infty(\O)},
\end{equation}
for all $n\in\N$, and therefore the iterations are well defined.

The plan of the article is the following. In Section~2 we introduce the notion 
of decreasing rearrangement and establish the equivalence between 
\fer{def.GFF} and \fer{def.GFV}
and its corresponding versions under this transformation. In addition, we 
show some examples on the relationship between an image and its relative rearrangement and histogram.
In Section~3, we prove some qualitative properties of the nonlinear iterative scheme,
among which its behavior as a shock filter. In Section~4, we describe the 
discretization of the continuous model and demonstrate with examples the denoising 
capabilities of the NF in comparison with the Bilateral and NLM filters, and 
its interpretation as a segmentation-like algorithm. In Section~5 we give our conclusions.

Finally, let us emphasize that the reformulation of the Neighborhood filter 
we propose produces the same filtered image than that obtained through the classical 
NF. However, there are two important advantages in our approach:
(i) the filtering process is performed through a one-dimensional
integral, reducing largely the algorithm complexity, and (ii) the rearranged version exposes the NF functioning to a deeper mathematical analysis.

\section{Neighborhood filters in terms of the decreasing rearrangement}

Let us denote by $\abs{E}$ the Lebesgue measure of any measurable set $E$.

For a Lebesgue measurable function $u:\O\to\R$, the function 
$q\in\R\to m_u(q) = \abs{\{\bx \in\O : u(\bx) >q\}}$ is called the \emph{distribution function} corresponding to $u$. 

Function $m_u$ is non-increasing and therefore admits a unique  generalized inverse, called the
\emph{decreasing rearrangement}. This inverse takes the usual pointwise meaning when 
the function $u$ has not flat regions, i.e. when $\abs{\{\bx \in\O : u(\bx) =q\}} =0$ for any $q\in\R$. In general, 
the decreasing rearrangement $u_*:[0,\abs{\O}]\to\R$ is given by:
\begin{equation*}
u_*(s) =\left\{
\begin{array}{ll}
 {\rm ess}\sup \{u(\bx): \bx \in \O \} & \qtext{if }s=0,\\
 \inf \{q \in \R : m_u(q) \leq s \}& \qtext{if } s\in (0,\abs{\O}),\\
 {\rm ess}\inf \{u(\bx): \bx \in \O \} & \qtext{if }s=\abs{\O}.
\end{array}\right.
 \end{equation*}
Notice that since $u_*$ is non-increasing in $[0,\abs{\O}]$, it is continuous but at most a countable subset of  
$[0,\abs{\O}]$. In particular, it is right-continuous for all $t\in (0,\abs{\O}]$.

The notion of rearrangement of a function is classical. We refer the reader to the textbook
\cite{Lieb2001} for the basic definitions and to the monograph \cite{Rakotoson2008} for a deeper insight into the subject.

The following equi-measurability property holds \cite[Corollary 1.1.1]{Rakotoson2008}. 
Let $F:\R\to\R_+$ be a Borel function. Then 
\begin{equation}
\label{prop.1}
 \int_\O F(u(\by))d\by = \int_0^{\abs{\O}} F(u_* (s))ds.
\end{equation}
In particular, $\nor{u}_{L^p(\O)}=\nor{u_*}_{L^p(0,\abs{\O})}$ for all $p\in[0,\infty]$.

We shall use the following notation for the level sets of $u$:
\begin{equation}
 \label{def.levelsets}
L_t(u)=\{\by \in \O : u(\by)=u_*(t) \},\qtext{for }t \in [0,\abs{\O}].
\end{equation}

The following theorem asserts that we may compute the Neighborhood filters in 
the one-dimensional space  $[0,\abs{\O}]$. 

\begin{theorem}
 \label{th.equivalence1}
Let $\O\subset\R^d$ be an open and bounded set, $d\geq 1$, and $\cK:\R\to\R_+$ be a Borel function. Let $u^{(0)}\in L^\infty(\O)$ be,
without loss of generality, non-negative and set $v_0 =u^{(0)}_*$. 

Then, for $m=0$ (resp. $m=n$), the iterative scheme \fer{def.GFF} (resp \fer{def.GFV}) 
 may be computed as, for $\bx \in L_t(u^{(0)})$ and $t\in [0,\abs{\O}]$, $u^{(n+1)}(\bx)=   v_{n+1}(t)$, with
 \begin{equation}
\label{def.NFstar}
  v_{n+1}(t) = \frac{1}{c_m(t)} \int_0^{\abs{\O}} \cK_h(v_{m}(t)-v_{m}(s))v_{n}(s)ds,
\end{equation}
  and $c_m(t)=\int_0^{\abs{\O}} \cK_h(v_{m}(t)-v_{m}(s))ds$.
 In addition, for $n\in \N$,
 \begin{equation}
\label{bound.iterstar}
 \nor{v_{n+1}}_{L^\infty(0,\abs{\O})}\leq \nor{v_n}_{L^\infty(0,\abs{\O})}.
\end{equation}
\end{theorem}
\noindent\emph{Proof. }
We start considering the one-step Neighborhood filter defined in \fer{def.NF}.
For each $\bx\in \O$ such that $u(\bx) <\infty$ (i.e. all $\bx\in\O$ but a subset of zero measure), we consider the Borel function $F:\R\to\R_+$ given by 
\begin{equation}
 \label{def.F}
F(w)=\cK_h(u(\bx)-w)w\qtext{if }w\geq 0,\quad F(w)=0\qtext{if }w<0. 
\end{equation}
 Then, \fer{prop.1} implies
\begin{equation*}
 \int_\O \cK_h(u(\bx)-u(\by))u(\by)d\by = \int_0^{\abs{\O}} \cK_h(u(\bx)-u_*(s))u_* (s))ds,
\end{equation*}
and
\begin{equation*}
 C(\bx)= \int_0^{\abs{\O}} \cK_h(u(\bx)-u_*(s))ds.
\end{equation*}
Since $\bx \in L_t(u^{(0)})$,
for some $t\in [0,\abs{\O}]$, we have   
\begin{align*}
 \int_0^{\abs{\O}} \cK_h(u(\bx)- & u_*(s)) u_* (s))ds  
  =\int_0^{\abs{\O}} \cK_h(u_*(t)-u_*(s))u_* (s))ds,
\end{align*}
and similarly for $C(\bx)$. 

We thus introduce the equivalent formulation to \fer{def.NF} given by, for $t\in [0,\abs{\O}]$
\begin{equation}
\label{def.onestep}
\NF^h u(\bx)=\NF^h u_* (t) \qtext{if}\quad \bx \in L_t(u) ,
\end{equation}
with
\begin{equation*}
 \NF^h u_* (t) = \frac{1}{c(t)} \int_0^{\abs{\O}} \cK_h(u_*(t)-u_*(s))u_* (s)ds,
\end{equation*}
and $c(t)=\int_0^{\abs{\O}} \cK_h(u_*(t)-u_*(s))ds$.

The extension of this transformation to the iterative schemes \fer{def.GFF} and \fer{def.GFV} ($m=0$ or $m=n$, respectively) 
is straightforward
 due to the invariance of the level sets structure with respect to the Neighborhood filtering, see \fer{def.onestep}. 
For both schemes, the step $n=1$ gives 
\begin{equation*}
u^{(1)}(\bx)= v_1(t):=\NF^h u_*^{(0)} (t) , 
\end{equation*}
for $\bx \in L_t(u^{(0)})$.
 In the case of the scheme with variable kernel \fer{def.GFV}, we may use the same 
function $F$ defined in \fer{def.F} to deduce that $u^{(2)}$ is constant in 
$L_t(u^{(0)})$ for each $t\in [0,\abs{\O}]$. An induction argument allows us to define
\begin{equation*}
u^{(n+1)}(\bx)=  v_{n+1}(t):=\NF^h v_n (t) , 
\end{equation*}
for $n\in\N$, and $\bx \in L_t(u^{(0)})$, which is  \fer{def.NFstar} with $m=n$.

For the fixed kernel scheme \fer{def.GFF} ($m=0$) we can not use $F$ in the step $n=1$ because 
the values of $u^{(0)}$ and $u^{(1)}$ inside the integral may be different on the level set $L_t(u^{(0)})$. 
However, these functions are still constant in each level set, 
implying the existence of a measurable function $f_1$ such that $f_1(u^{(1)}(\bx))=u^{(0)}(\bx)$
for $\bx\in L_t(u^{(0)})$, for $t\in [0,\abs{\O}]$. Therefore, we may repeat the above argument with function $F$ replaced 
by $F_1(w)=\cK_h(u^{(0)}(\bx)-f_1(w))w$ if $w\geq0$ and $F_1(w)=0$ if $w<0$, thus obtaining 
\begin{align*}
 u^{(2)}(\bx)= & \NF^h u^{(1)}(\bx)  
 =  \frac{1}{C_1(\bx)}
 \int_0^{\abs{\O}} \cK_h(u^{(0)}(\bx)-f_1(v_1(s))v_1(s)ds \\
 = & \frac{1}{C_1(\bx)}
 \int_0^{\abs{\O}} \cK_h(u^{(0)}(\bx)-u^{(0)}_*(s))v_1(s)ds.
\end{align*}
Reasoning in a similar way for  $C_1(\bx)$, and recalling the definition of $v_0$ we obtain for 
 $\bx \in L_t(u^{(0)})$,
\begin{align*}
 u^{(2)}(\bx)=v_2(t):=  \frac{1}{c_1(t)}
 \int_0^{\abs{\O}} \cK_h(v_0(t)-v_0(s))v_1(s)ds.
\end{align*}
Then \fer{def.NFstar} for $m=0$. follows from an inductive argument.
 
Finally, using \fer{bound.iter} and the equi-measurability property \fer{prop.1} we obtain
\fer{bound.iterstar} for all $n\in\N$.
$\Box$
\begin{figure*}[ht]
\centering
 \subfigure%
  {\includegraphics[width=4cm,height=4cm]{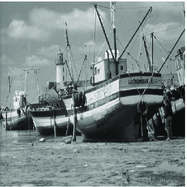}}
 \subfigure%
 {\includegraphics[width=4cm,height=4cm]{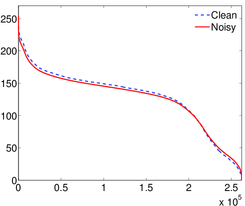}}
 \subfigure%
 {\includegraphics[width=4cm,height=4cm]{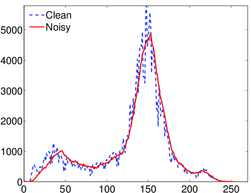}} \\
 \subfigure%
 {\includegraphics[width=4cm,height=4cm]{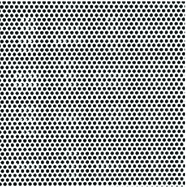}}
 \subfigure%
 {\includegraphics[width=4cm,height=4cm]{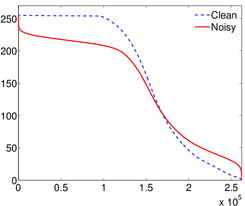}}
 \subfigure%
 {\includegraphics[width=4cm,height=4cm]{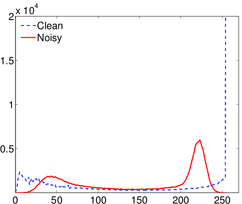}}  \\
 \subfigure%
 {\includegraphics[width=4cm,height=4cm]{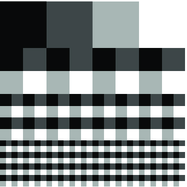}}
 \subfigure%
 {\includegraphics[width=4cm,height=4cm]{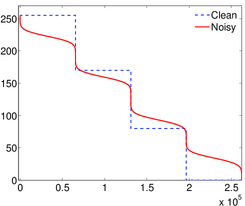}}
 \subfigure%
 {\includegraphics[width=4cm,height=4cm]{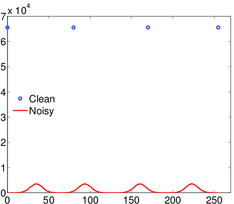}} 
  \caption{Test images \emph{Boat}, \emph{Texture} and \emph{Squares}. 
  Second and third columns show the decreasing rearrangement and the histogram, respectively, of the original images (clean) and their counterparts obtained after the addition of a Gaussian noise with SNR$=10$.}
\label{fig1}
\end{figure*}

\subsection{Examples}

We show three examples of the decreasing rearrangement for clean and noisy images, all of them 
quantized in the usual interval $[0,255]$. 

We have chosen as test images a natural image, \emph{Boat}, 
a texture image, \emph{Texture}, and a synthetic image, \emph{Squares}.
The first two images are taken from the data base of the 
Signal and Image Processing Institute,  University of Southern California
({\tt boat.512.tiff} of Vol.~3 and  {\tt 1.5.02.tiff} of Vol.~1, 
respectively), see \cite{Collins1998,Kwan1999}, while the third is a synthetic image constructed with four gray levels 
($0,~85,~170$ and $255$) and such that its four level sets have the same measure.

This choice is motivated by the bad (resp. good) performance of the NF for 
uniform (resp. extreme) distribution of gray levels mass. These distributions are plotted in Fig.~\ref{fig1}. 

A gray levels mass uniformly distributed image has a straight line as decreasing rearrangement, or a constant, as histogram. In Fig.~\ref{fig1}, we observe that the decreasing rearrangement  of the Boat, is closer to a straight line than that of the Texture, which is still  \emph{continuous}. The choice of the synthetic image is motivated by its extreme behavior: a piece-wise constant decreasing rearrangement.

We have added a Gaussian white noise of $\text{SNR}=10$ to the test images 
according to the noise measure $\text{SNR}=\sigma(u)/\sigma(\nu)$,
where $\sigma$ is the empirical standard deviation, $u$ is the original image, and $\nu$ is the noise. 

We may observe in  Fig.~\ref{fig1} that the main consequence of noise addition on the decreasing rearrangement is its smoothening towards a straight line. Of course, the effect is stronger in 
far from uniformly distributed images, like the Squares.

Finally, observe the connection between points of local maximum or minimum
for the histogram and inflexion points for the decreasing rearrangement.
This is the base for justifying the use of the NF as a segmentation algorithm.

\section{Properties of the nonlinear varying-kernel iterative scheme}

For the differential analysis we carry out in this section we have to assume regularity properties on the decreasing rearrangement of the given image.

\begin{figure}[t]
\centering
  \subfigure
 {\includegraphics[width=5cm,height=4cm]{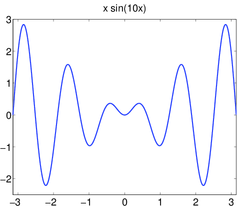}}
 \hspace{1cm}
 \subfigure
 {\includegraphics[width=5cm,height=4cm]{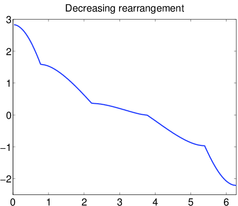}}
 \caption{{\small Graphic of function $f(x)=x\sin(10x)$ (top panel) and that of its decreasing rearrangement, $f_*$ (bottom panel). 
 Although $f\in C^\infty (-\pi,\pi)$, $f_*$ is not even once continuously differentiable in $(0,2\pi)$. }} 
\label{rem.fig}
\end{figure}

In general, if $u\in W^{1,q}(\O)$, with $q>d$, and $\O$ is open, bounded and connected, then  $u_* \in W^{1,p}(0,\abs{\O})$ for any $p<2$, see 
\cite[Th. 3.3.2]{Rakotoson2008}.  
However, not much more than this is expected, and even for 
$C^\infty (\O)$ functions, their decreasing rearrangement may be non globally 
differentiable, see Fig.~\ref{rem.fig}.

The following theorem asserts that the monotonicity property of the decreasing rearrangement of the initial image
is conserved along all the iterations. 
In addition, for some type of kernels among which the Gaussian is included, 
the measure of flat regions does not change in the iterative procedure. We also obtain a
condition in terms of the window size, $h$, ensuring that the steady state is a constant.

\begin{theorem}
\label{th.derivada}
 Let $v_0= u^{(0)}_* \in W^{1,p}(0,\abs{\O})$ for some $p\geq 1$.
 Let  $\cK\in W^{1,\infty}_{\text{loc}}(\R)$,  with $\cK (\xi)\geq 0$ for all $\xi\in\R$, 
 and $\cK(\xi) > 0$ for all $\xi$ in a neighborhood of $0$. 

Let $v_{n+1}$ be given by the iterative scheme  \fer{def.NFstar}, for $n\in\N$ and $m=n$.
 Then we have, for all $n\in\N$,
 \begin{enumerate}
  \item[(i)] $v_{n+1}\in W^{1,p}(0,\abs{\O})$,  and if $v'_0(t)=0$ then $v'_{n+1}(t)=0$.
  
    \item[(ii)] Let $R_1:=(\xi_1-\xi_2)\big(\cK'(\xi -\xi_1)\cK(\xi -\xi_2)-\cK'(\xi -\xi_2)\cK(\xi -\xi_1)\big)$ and
    assume 
\begin{equation}
\label{cond.k}
R_1\geq 0 \text{ for all }\xi,\xi_1,\xi_2 \in\R.
\end{equation}
Then $v'_{n+1} \leq 0$ a.e. in $(0,\abs{\O})$. In addition, if $v'_0(t)<0$ and the inequality in 
\fer{cond.k} is strict then $v'_{n+1}(t)<0$.
  
  \item[(iii)] Let $\phi:\R_+\to\R_+$ be a continuous function 
  of $h$ such that $\lim_{h\to \infty} \phi(h)=0$. 
  Assume that $\cK >0$ in $\R$, and let 
  \begin{equation*}
  \label{def.R2}
 R_2= (\xi_1-\xi_2)\Big(\frac{\cK'(\xi -\xi_1)}{\cK(\xi -\xi_1)}-\frac{\cK'(\xi -\xi_2)}{\cK(\xi -\xi_2)}\Big).
\end{equation*}
Assume
  \begin{equation}
   \label{cond.k2}
   R_2 \leq \phi(h)(\xi_1-\xi_2)^q \quad \forall h\geq0,\quad \forall \xi,\xi_1,\xi_2 \in\R,
  \end{equation}
  for some $q\geq 1$.
  Then, if $h$ is large enough the sequence $v_n$ converges uniformly to a constant.
   \end{enumerate}
\end{theorem}
  
\begin{remark}
With the additional assumption of
$\cK$ being symmetric, 
condition \fer{cond.k} imposes $\cK(\xi_2) \geq \cK(\xi_1)$
if $\abs{\xi_1}\geq \abs{\xi_2}$, that is, it must be a decaying kernel. 

Observe that the Gaussian kernel $\cK(s)=\text{e}^{-s^2}$ satisfies all the assumptions 
of Theorem~\ref{th.derivada}. In particular, condition \fer{cond.k2} is satisfied for $\phi(h)=1/h^2$ and $q=2$.

Functions $R_1$ and $R_2$ are related in the following way: $R_1=\cK(\xi-\xi_1)\cK(\xi-\xi_2)R_2$. Thus, if 
$\cK>0$ in $\R$ then condition \fer{cond.k} is equivalent to $R_2 \geq 0$. 
\end{remark}
    
\emph{Proof of Theorem~\ref{th.derivada}.}
Differentiating  \fer{def.NFstar}, for $m=n$, with respect to $t$ we obtain
\begin{align}
  v'_{n+1}(t)  = &  \label{deri} 
   \frac{1}{c_n^2(t)}v'_{n} (t)
 \Big(c_n(t) \int_0^{\abs{\O}}\cK_h'(v_{n}(t)-v_{n}(s)) v_{n}(s) ds \\
 & - \int_0^{\abs{\O}} \cK_h(v_{n}(t)-v_{n}(s))v_{n}(s)ds 
   \int_0^{\abs{\O}}\cK_h'(v_{n}(t)-v_{n}(s)) ds \Big). \nonumber
\end{align}
Let us consider the case $n=0$. By assumption, $v_0 \in W^{1,p}(0,\abs{\O})$. 
On one hand, since $\cK$ is continuous and positive in a neighborhood of $0$, and $v_0'$ is bounded in $L^1(0,\abs{\O})$, we deduce
$c_0(t) \geq \hat c \int_{\R} \cK_h(\xi) d\xi$, for some positive constant $\hat c$. 

On the other hand, the regularity assumed on $v_0$ also implies $v_0\in L^\infty(0,\abs{\O})$. Using these properties together with 
the Lipschitz continuity of $\cK$ we obtain that also $v_1\in W^{1,p}(0,\abs{\O})$. Then we proceed recursively to deduce $v_{n+1}\in W^{1,p}(0,\abs{\O})$
for all $n\in\N$, and thus (i) follows.

To check the sign of $v'_{n+1}$, let us simplify the notation introducing, for fixed $t\in(0,\abs{\O})$,
\begin{align*}
 & f(s)=\cK_h'(v_{n}(t)-v_{n}(s)),\quad
 g(s)=v_{n}(s),\\
 & h(s)=\cK_h(v_{n}(t)-v_{n}(s)).
\end{align*}
Then $v'_{n+1} \leq 0$ follows from \fer{deri} if we prove
\begin{align*}
 \int_0^{\abs{\O}} f(s)ds &\int_0^{\abs{\O}} g(z)h(z)dz \leq 
  \int_0^{\abs{\O}} h(s)ds \int_0^{\abs{\O}} f(z)g(z)dz,
\end{align*}
which is equivalent to $\alpha\geq 0$, with 
\begin{equation}
\label{def.alpha}
 \alpha=\int_0^{\abs{\O}}\int_0^{\abs{\O}}g(z)\big(f(z)h(s)-f(s)h(z)\big)dsdz.
\end{equation}
Interchanging the dummy variables, we get after addition of the corresponding 
identities
\begin{equation*}
\label{alpha.2}
2 \alpha=\int_0^{\abs{\O}}\int_0^{\abs{\O}}(g(z)-g(s))\big(f(z)h(s)-f(s)h(z)\big)dsdz.
\end{equation*}
We then use assumption \fer{cond.k} to deduce $\alpha\geq 0$. 
The second part of  (ii) follows from similar arguments.

We, finally, prove (iii). Using \fer{deri} and the definition \fer{def.alpha}  we have
\begin{equation*}
 v_{n+1}'(t)=\frac{\alpha}{c_n^2(t)} v_{n}'(t).
\end{equation*}
Since $\cK>0$, we may rewrite $\alpha$ as 
\begin{equation*}
\alpha  =\int_0^{\abs{\O}}\int_0^{\abs{\O}}g(z)h(s)h(z) \Big(\frac{f(z)}{h(z)}-\frac{f(s)}{h(s)}\Big)dsdz.
\end{equation*}
Using again the interchange of dummy variables and condition \fer{cond.k2} we obtain
\begin{equation*}
 \frac{\alpha}{c_n^2(t)} \leq  \phi(h) \nor{v_n}_{L^\infty}^q.
\end{equation*}
Therefore, for $h$ large enough (depending on the shape of function $\phi$) we have $\abs{v_{n+1}'(t)}\leq c \abs{v_{n}'(t)}$, with
$c<1$. The result follows. $\Box$

\begin{corollary}
\label{cor.reg}
 Let $v_0= u^{(0)}_* \in C^m([0,\abs{\O}])$ and $\cK\in C^m(\R)$, with $\cK >0 $. Then 
 $v_{n+1} \in C^m([0,\abs{\O}])$ for all $n\in\N$.
\end{corollary}

\noindent\emph{Proof. } From formula \fer{deri} we see that the $m-$order derivative of $v_{n+1}$
is given in terms of a quotient with a non-vanishing denominator and a numerator expressed as a composition of continuous functions, given in terms of the derivatives of order up to $m$ of $v_n$ and $\cK$. $\Box$ 

Although an easy consequence of Theorems~\ref{th.equivalence1} and \ref{th.derivada}, the following result neatly exposes the functioning of the Neighborhood filter. We recall that 
the mapping $g:\R\to\R$ is a \emph{contrast change} if $g$ is strictly increasing and continuous.
\begin{corollary}
\label{cor.contrast}
 Assume the conditions of Theorem~\ref{th.derivada}, with the exception of condition (iii), 
 and suppose that $v_0'(t)<0$ for all $t\in(0,\abs{\O})$.
 Then, there exists a contrast change, $g$ such that $u^{(n+1)}=g(u^{(0)})$ in $\O$, where $u^{(n+1)}$ is the $(n+1)-$th iteration of the nonlinear iterated Neighborhood filter \fer{def.GFV}. 
\end{corollary}
\noindent\emph{Proof. } 
Since, by assumption, $v_0\in W^{1,1}(0,\abs{\O})\subset C([0,\abs{\O}])$, and  $v_0'(t)<0$ for $t\in(0,\abs{\O})$, the corresponding distribution function of $v_0$, $m_{v_0}$
is continuous and invertible in $(0,\abs{\O})$ and, actually, it is the inverse of $v_0$. 
Notice that  $m_{v_0}$ coincides with the distribution function of 
$ u^{(0)}_*$, since $v_0= u^{(0)}_*$.
We define the function 
\[
  g(q)=v_{n+1}(m_{v_0}(q)), \qtext{for } q\in [\min{v_0},\max{v_0}].
\]
By Theorem~\ref{th.derivada}, $v_{n+1}\in W^{1,1}(0,\abs{\O})\subset C([0,\abs{\O}])$, and $v_{n+1}'(t)<0$, implying that $g$ is continuous and increasing (composition of two continuous and decreasing functions). 

According to Theorem~\ref{th.equivalence1}, the structure of level sets is invariant under the NF, i.e. the level lines of $u^{(n+1)}$ are the same as those 
of $u^{(0)}$. In addition, the values of $u^{(n+1)}$ on the level lines are given by \fer{def.NFstar}. Therefore,  for 
$\bx \in L_s(u^{(0)})$, for all $s\in[0,\abs{\O}]$, we have
\[
 g(u^{(0)}(\bx))=g(v_0(s))=v_{n+1}(s)=u^{(n+1)}(\bx).
\]
$\Box$

In the following theorem we establish a correspondence between the nonlocal diffusion 
scheme \fer{def.NFstar} and local diffusion. This is a fundamental ingredient for the properties deduced later, which can not be 
directly deduced from the N-dimensional model.

Indeed, since $v_n$ is non-increasing for all $n\in\N$ (Theorem~\ref{th.derivada}), we have that the values selected by the nonlocal kernel $\cK_h(v_n(t)-v_n(s))$ are related to the independent variables through, for instance, Taylor's expansion. 

For example, if $u_*$ is smooth, we may approximate
\begin{equation*}
\label{approxi}
 \int_\O \text{e}^{-\frac{(v_n(t)-v_n(s))^2}{h^2}}v_n(s)ds \approx
 \int_\O \text{e}^{-\frac{((t- s)v_n'(t))^2}{h^2}}v_n(s)ds.
\end{equation*}
Therefore, the size of the support of the cut-off function approximated by the Gaussian is related to 
the size of $v_n'$. If $v_n'$ is large, the size is small and the diffusion only occurs in a small interval
around $v_n(t)$. The opposite effect holds if $v_n'$ is small.

Although more general assumptions on $\cK$ may be prescribed, see Remark~\ref{remth.pde}, we estate this result for the Gaussian kernel, for clarity. We also ask for further regularity on $u_*$.

\begin{theorem}
\label{th.pde}
Let $v_0=u_* \in C^3([0,\abs{\O}])$ be such that $v_0'<0$ in $[0,\abs{\O}]$. Let 
$ \cK(\xi)=\text{e}^{-\xi^2}$.
Then, for all $t\in(0,\abs{\O})$, there exist positive constants $\alpha_1,\alpha_2$ independent of $h$ such that
\begin{align}
\label{app.NFstar}
v_{n+1}(t) = v_{n}(t)&+\alpha_1 \tilde k_h(t)v_{n} ' (t) \big( h+O(h^{3/2}) \big) 
 - \alpha_2 \frac{v_{n} '' (t)}{(v_{n} ' (t))^2} h^2 + O(h^{5/2}),
\end{align}
with 
\begin{equation}
\label{def.ktilde}
 \tilde k_h(t)= \frac{\cK_h(v_n(t)-v_n(\abs{\O}))}{v_n'(\abs{\O})}-
 \frac{\cK_h(v_n(t)-v_n(0))}{v_n'(0)},
\end{equation}
and with $\alpha_1\approx 1/\sqrt{\pi}$, and $\alpha_2\approx 1$.
\end{theorem}

 There are two interesting effects captured by \fer{app.NFstar}:
 \begin{enumerate}
  \item The border effect (loss of contrast). Function $\tilde k_h$ is \emph{active} only when $t$ is close to the boundaries, $t\approx 0$ and $t\approx \abs{\O}$. For 
   $t \approx 0$  the term $\tilde k_h(t)v_{n} ' (t) <0 $ contributes to the decrease of 
  the largest values of $v_{n+1}$ while for $t\approx \abs{\O}$ the opposite effect takes place. Therefore, 
  this term tends to flatten $v_{n+1}$, i.e. induces a loss of contrast.
  
  \item The term $-\frac{h^2}{2} \frac{v_{n} '' (t)}{(v_{n} ' (t))^2}$ is anti-diffusive, inducing large 
  gradients of the iterated functions in a neighborhood of the inflexion points. In this sense, the scheme \fer{app.NFstar}
  is related to the shock filter introduced by Alvarez and Mazorra \cite{Alvarez1994} 
  \begin{equation}
  \label{alvarez}
   u_t+F(G_t u_{xx},G_t u_{x})u_x=0,
  \end{equation}
where $G_t$ is a smoothing kernel and function $F$ satisfies $F(p,q)pq\geq 0$ for any $p,q\in\R$. Indeed, neglecting the 
border and the lower order terms, and defining $F(p,q)=p/q^3$, we obtain from \fer{app.NFstar}
\begin{equation*}
 v_{n+1}(t) - v_{n}(t)+  \alpha_2 h^2 F(v_{n} '' (t),v_{n} ' (t)) v_{n} ' (t)=0 ,
\end{equation*}
which may be regarded as a time discretization of \fer{alvarez}.
 \end{enumerate}

\emph{Proof of Theorem~\ref{th.pde}. }
We may rewrite the iterative scheme  \fer{def.NFstar}, for $n\in\N$ and $m=n$
as
\begin{align}
\label{th2.1}
 v_{n+1}&(t)-v_n(t)= \\ 
 & \frac{1}{c_n(t)}\int_0^{\abs{\O}} \cK_h(v_n(t)-v_n(s))(v_n(s)-v_n(t))ds.\nonumber
\end{align}
Due to (3) of Theorem~\ref{th.derivada} we have $v_{n}'<0$ in $(0,\abs{\O})$, and due to \fer{bound.iterstar}, $v_n(0,\abs{\O})\subset v_0(0,\abs{\O})$.
Let us denote the  inverse of $v_{n}$ by $v_n^{-1}$.
Using the change of variable $s=v_n^{-1}(q)$ and writing
$t=v_n^{-1}(z)$, we obtain from \fer{th2.1}
\begin{equation}
\label{th2.3}
 v_{n+1}(t)-v_n(t)=\frac{I_1(z)}{I_2(z)},
\end{equation}
with 
\begin{align*}
 & I_1(z)= \int_{v_n(\abs{\O})}^{v_n(0)}  \cK_h(z-q)(q-z)\frac{dq}{v_{n} ' (v_n^{-1}(q))},\\ 
 & I_2(z)=\int_{v_n(\abs{\O})}^{v_n(0)} \cK_h(z-q)\frac{dq}{v_{n} ' (v_n^{-1}(q))}.
\end{align*}
Using the explicit form of $\cK$ and integrating by parts, we obtain
\begin{align}
\label{th2.i1}
 I_1(z)=& \frac{h^2}{2}\Big( \tilde k_h (v_n^{-1}(z)) 
  -\int_{v_n(\abs{\O})}^{v_n(0)} \cK_h(z-q)\frac{v_{n} '' (v_n^{-1}(q))}{(v_{n} ' (v_n^{-1}(q)))^3}dq\Big), 
\end{align}
with $\tilde k_h$ given by \fer{def.ktilde}.

By assumption, functions
\begin{equation*}
 f(q)=\frac{v_{n} '' (v_n^{-1}(q))}{(v_{n} ' (v_n^{-1}(q)))^3} \qtext{and}\quad g(q)=\frac{1}{v_{n} ' (v_n^{-1}(q))}
\end{equation*}
are bounded in $[v_n(\abs{\O}),v_n(0)]$ and by Corollary~\ref{cor.reg} they are also continuously differentiable in $(v_n(\abs{\O}),v_n(0))$.

Consider the interval $J_h=\{q: \abs{z-q}<\sqrt{h}\}$.
By well known properties of the Gaussian kernel, we have
\begin{equation}
 \label{gauss.1}
 \kappa(h): = \int_{J_h} \cK_h(z-q) dq <  \int_\R \cK_h(q) dq = h\sqrt{\pi},
\end{equation}
and
\begin{equation}
 \label{gauss.2}
\cK_h(z-q)\leq  \text{e}^{-1/h} \quad\text{if}\quad  q\in J_h^C=\{q:\abs{z-q}\geq\sqrt{h}\}.
\end{equation}
 In particular, from \fer{gauss.2} we get 
\begin{equation}
\label{th2.4}
 \left| \int_{J_h^C} \cK_h(z-q)f(q)dq \right| < O(h^\alpha)\qtext{for any }\alpha >0.
\end{equation}
Taylor's formula implies 
\begin{align*}
 \int_{v_n(\abs{\O})}^{v_n(0)}   \cK_h& (z-q)  f(q)dq  
 =  \int_{J_h} \cK_h(z-q) (f(z) +O(\sqrt{h})) dq 
 + \int_{J_h^C} \cK_h(z-q)f(q)dq .
\end{align*}
Therefore, from \fer{th2.i1}, \fer{gauss.1} and \fer{th2.4} we deduce
\begin{align*}
I_1(z)=\frac{h^2}{2} \Big( \tilde k(v_n^{-1}(z)) -
\frac{v_{n} '' (v_n^{-1}(z))}{(v_{n} ' (v_n^{-1}(z)))^3} \kappa(h) + O(h^{3/2})\Big).
\end{align*}
Similarly,
\begin{align*}
 I_2(z)& =\int_{v_n(\abs{\O})}^{v_n(0)}  \cK_h(z-q)g(q)dq 
= \int_{J_h} \cK_h(z-q) (g(z)+O(\sqrt{h})) dq 
 + \int_{J_h^C} \cK_h(z-q)g(q)dq \\
&  =\frac{1}{v_{n} ' (v_n^{-1}(z))} \kappa(h) + O(h^{3/2}).
\end{align*}
Then, 
the result follows from \fer{th2.3} substituting $z$ by $v_n(t)$. $\Box$

\begin{remark}
 \label{remth.pde}
 Theorem~\ref{th.pde} may be extended to Lipschitz continuous decaying kernels satisfying
 the growth condition 
 \begin{equation}
  \label{ext.kernel}
   K(s)\leq \frac{k_0}{1+\abs{s}^p},\qtext{for some }p>1.
\end{equation}
In such case, the higher order terms in formula \fer{app.NFstar} must be replaced by 
$O(h^{\alpha+1})$ and $O(h^{\alpha +2})$, respectively, with $\alpha=(p-1)/(p+1)$, and 
$\alpha_1,~\alpha_2$ are just some positive constants. In addition, function $\tilde k_h$
given by \fer{def.ktilde} is replaced by some continuous function related to a primitive of $sK_h(s)$, and still inducing the border effect commented after the statement of the theorem. This primitive plays the same role as $-h^2K_h(s)/2$ (the primitive when $K$ is a Gaussian kernel) in formulas \fer{th2.i1}, 
\fer{gauss.1} and \fer{gauss.2}, from where condition \fer{ext.kernel} arises.
\end{remark}

\section{Discretization and numerical examples}

For computing the iterated Neighborhood filter \fer{def.GFV} of a function through its 
decreasing rearrangement version \fer{def.NFstar}, we assume that the initial image, $u^{(0)}$ is quantized in some range, e. g. $[0,255]$, and compute its decreasing rearrangement $v_0=u^{(0)}_*$ as the inverse of the distribution function $m_{u^{(0)}}$.
Then, the iterations are performed by computing the integrals involved in the filter by a simple middle point formula, i.e.  by assuming a constant-wise interpolation of the discrete image. 

Only two parameters must be fixed in advance, the \emph{length} of the kernel window, $h$, and a tolerance for the stopping criterium or, alternatively, the number of filtering iterations.

When the iterations are stopped at iteration, say, $n+1$, we recover the output image by using the formula provided in Theorem~\ref{th.equivalence1},
\[
  u^{(n+1)}(\bx)=   v_{n+1}(t) \qtext{for }\bx \in L_t(u^{(0)})\qtext{and } t\in [0,\abs{\O}], 
\]
where $L_t(u^{(0)})$ stands for the level sets of $u^{(0)}$, see \fer{def.levelsets}.
Recall that the level sets structure of $u^{(n)}$, for $n=0,1,\ldots,$ is invariant.

We used a stopping criterium based on the variational approach of the NF given by 
Kindermann et al. \cite{Kindermann2005}. In particular, the authors formally show that the 
critical points of the functional 
\begin{equation*}
 J(u)=\int_{\O\times\O} g \Big(\frac{(u(\bx)-u(\by))^2}{h^2}\Big)d\bx d\by,
\end{equation*}
for $g(s)=\int_0^s\cK_h(\sqrt{t}) dt$ , coincide with the fixed points of the Neighborhood filter.
The gradient descent scheme associated to the minimization of $J$ is just
the iterated Neighborhood filter \fer{def.GFV}, and thus the relative difference of 
the decreasing sequence $J(u^{(n+1)})$ between successive iterations may be used as a stopping criterium. In fact, using
the equi-measurability property \fer{prop.1} we readily deduce
\begin{equation*}
 J(u)=J_*(u_*):=\int_0^{\abs{\O}}\int_0^{\abs{\O}} g \Big(\frac{(u_*(s)-u_*(t))^2}{h^2}\Big)ds dt,
\end{equation*}
which is the actual form of the functional we use for the stopping criterium.

Let us mention that in \cite{Kindermann2005} the authors show that the functional
$J$ is not convex, in general, and therefore the existence and uniqueness of a global minimum for $J$ may not be deduced from the standard theory.

Finally, let us stress that in discrete computations the analytical results obtained in 
Section~3 are not always observed. 
The reason is, of course, that some of the assumptions are not fulfilled 
in the discrete framework. Importantly, those referring to the unbounded  
support of the kernel $\cK$, or to the regularity of $u_*$.

For example, for the Gaussian kernel used in our numerical experiments, 
it is proven in Theorem~\ref{th.derivada} that if $v_0'(t)<0$ then $v_n'(t)<0$ for all $n$. However, as it may be seen, for instance, in Fig.~\ref{fig2} (fourth row, second column), this property is
violated in the discrete framework. Nevertheless, the weaker result $v_n'(t)\leq 0$ is always observed in the experiments

\subsection{ Numerical examples for denoising}

In the first set of experiments we used the Neighborhood filter  for denoising porpouses, 
and compare it with other related filters: the Bilateral filter,
\begin{equation*}
 \BF^{h,\rho} u (\bx)=\frac{1}{C(\bx)}\int_\O \textrm{e}^{-\frac{\abs{u(\bx)-u(\by)}^2 }{h^2}}\textrm{e}^{-\frac{\abs{\bx-\by}^2 }{\rho^2}} u(\by)d\by,
\end{equation*}
where $h$ and $\rho$ are positive constants, and \[C(\bx)=\int_\O \exp\left(-\abs{u(\bx)-u(\by)}^2) h^{-2}\right)\exp\left(-\abs{\bx-\by}^2\rho^{-2}\right) d\by, \] 
and the Nonlocal Means filter,
\begin{equation*}
 \NL^{h,\rho} u (\bx)=\frac{1}{C(\bx)}\int_\O \textrm{e}^{-\frac{G_\rho * \abs{u(\bx+\cdot)-u(\by+\cdot)}^2 (0)}{h^2}} u(\by)d\by,
\end{equation*}
where $h>0$, $G_\rho$ is a Gaussian kernel of standard deviation $\rho>0$ and  \[C(\bx)=\int_\O \exp\left(-G_\rho * \abs{u(\bx+\cdot)-u(\by+\cdot)}^2 (0)) h^{-2}\right) d\by.\]

Since the usual version of the NF, given by \fer{def.GFV}, and the version introduced in this article, expressed through the decreasing rearrangement by \fer{def.NFstar}, are equivalent, there is no need of comparison between them.

The denoising properties of these three filters are well known, and a thoroughfull comparison among them (and among other filters) is given in \cite{Buades2010}. Here, we are not so interested in deciding which is the best performing denoising algorithm than in analyzing their behavior with respect to the histogram and the decreasing rearrangement redistributions.

We applied the filters on the test images given in the Introduction, see Fig.~\ref{fig1},
corrupted with an additive Gaussian white noise of $\text{SNR}=10$, 
according to the noise measure $\text{SNR}=\sigma(u)/\sigma(\nu)$,
where $\sigma$ is the empirical standard deviation, $u$ is the original image, and $\nu$ is the noise.

\begin{figure}[ht]
\centering
 {\includegraphics[width=4cm,height=4cm]{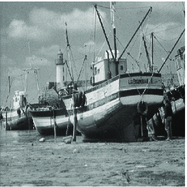}}
 {\includegraphics[width=4cm,height=4cm]{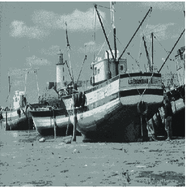}}
 {\includegraphics[width=4cm,height=4cm]{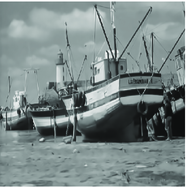}}
  {\includegraphics[width=4cm,height=4cm]{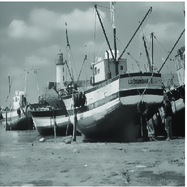}}\\
  {\includegraphics[width=4cm,height=4cm]{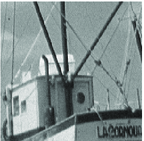}}
 {\includegraphics[width=4cm,height=4cm]{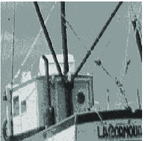}} 
 {\includegraphics[width=4cm,height=4cm]{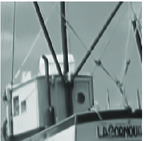}} 
  {\includegraphics[width=4cm,height=4cm]{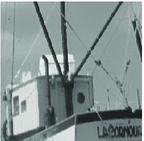}}\\
 \hspace{4cm}
 {\includegraphics[width=4cm,height=4cm]{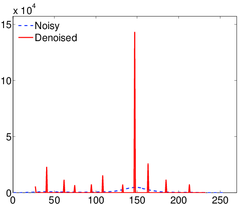}}
  {\includegraphics[width=4cm,height=4cm]{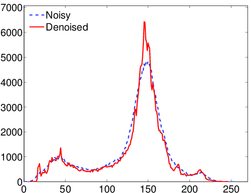}}
   {\includegraphics[width=4cm,height=4cm]{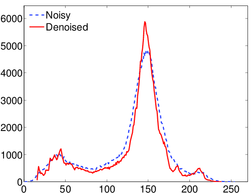}}\\
    \hspace{4cm}
 {\includegraphics[width=4cm,height=4cm]{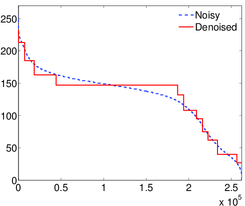}} 
 {\includegraphics[width=4cm,height=4cm]{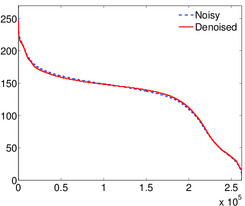}} 
 {\includegraphics[width=4cm,height=4cm]{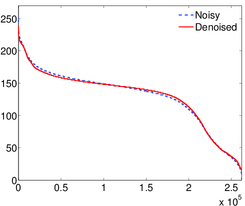}}  \\
  {\includegraphics[width=4cm,height=4cm]{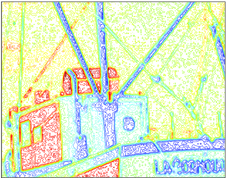}}
 {\includegraphics[width=4cm,height=4cm]{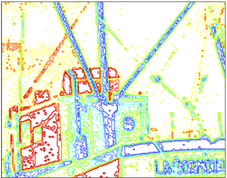}} 
  {\includegraphics[width=4cm,height=4cm]{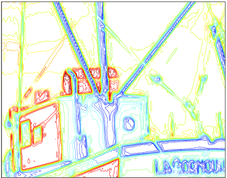}}
 {\includegraphics[width=4cm,height=4cm]{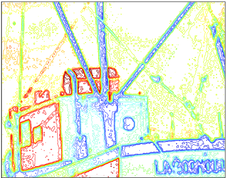}} \\
  \caption{{ Denoising experiment. Columns: noisy image, Neighborhood, NLM, and Bilateral filters. 
  Rows: noisy image, detail of the image, histograms, decreasing rearrangements, and 
  level curves of image details shown in row 2. }} 
\label{fig2}
\end{figure}

\begin{figure*}[ht]
\centering
 {\includegraphics[width=4cm,height=4cm]{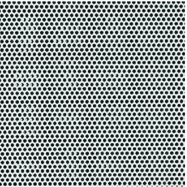}}
 {\includegraphics[width=4cm,height=4cm]{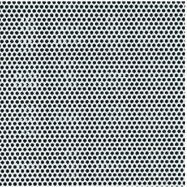}}
 {\includegraphics[width=4cm,height=4cm]{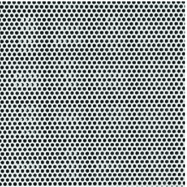}}
  {\includegraphics[width=4cm,height=4cm]{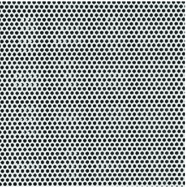}}\\
  {\includegraphics[width=4cm,height=4cm]{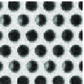}}
 {\includegraphics[width=4cm,height=4cm]{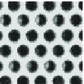}} 
 {\includegraphics[width=4cm,height=4cm]{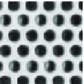}} 
  {\includegraphics[width=4cm,height=4cm]{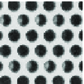}}\\
 \hspace{4cm}
 {\includegraphics[width=4cm,height=4cm]{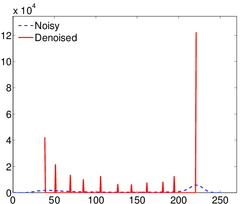}}
  {\includegraphics[width=4cm,height=4cm]{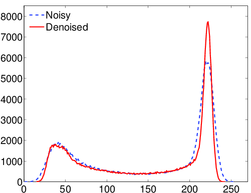}}
   {\includegraphics[width=4cm,height=4cm]{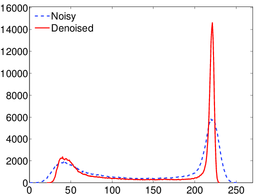}}\\
    \hspace{4cm}
 {\includegraphics[width=4cm,height=4cm]{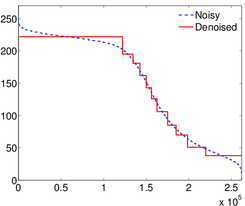}} 
 {\includegraphics[width=4cm,height=4cm]{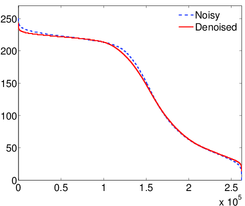}} 
 {\includegraphics[width=4cm,height=4cm]{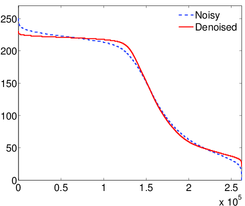}}  \\
  {\includegraphics[width=4cm,height=4cm]{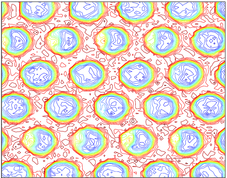}}
 {\includegraphics[width=4cm,height=4cm]{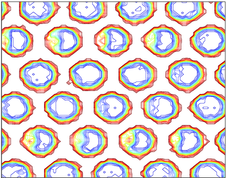}} 
  {\includegraphics[width=4cm,height=4cm]{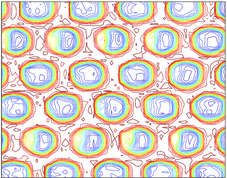}}
 {\includegraphics[width=4cm,height=4cm]{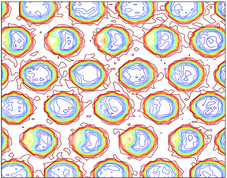}} \\
  \caption{{Denoising experiment. Columns: noisy image, Neighborhood, NLM, and Bilateral filters. 
  Rows: noisy image, detail of the image, histograms, decreasing rearrangements, and 
  level curves of image details shown in row 2.}} 
\label{fig3}
\end{figure*}

\begin{figure*}[ht]
\centering
 {\includegraphics[width=4cm,height=4cm]{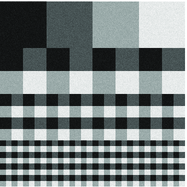}}
 {\includegraphics[width=4cm,height=4cm]{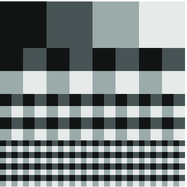}}
 {\includegraphics[width=4cm,height=4cm]{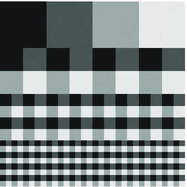}}
  {\includegraphics[width=4cm,height=4cm]{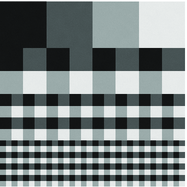}}\\
  {\includegraphics[width=4cm,height=4cm]{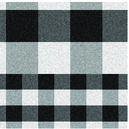}}
 {\includegraphics[width=4cm,height=4cm]{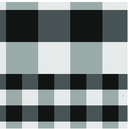}} 
 {\includegraphics[width=4cm,height=4cm]{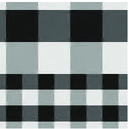}} 
  {\includegraphics[width=4cm,height=4cm]{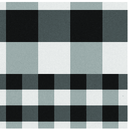}}\\
 \hspace{4cm}
 {\includegraphics[width=4cm,height=4cm]{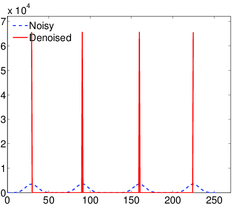}}
  {\includegraphics[width=4cm,height=4cm]{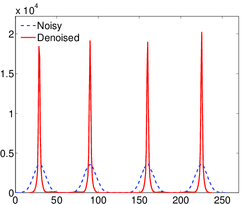}}
   {\includegraphics[width=4cm,height=4cm]{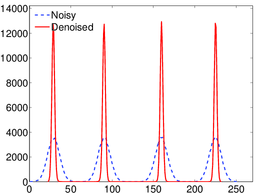}}\\
    \hspace{4cm}
 {\includegraphics[width=4cm,height=4cm]{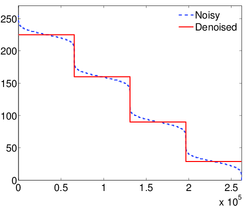}} 
 {\includegraphics[width=4cm,height=4cm]{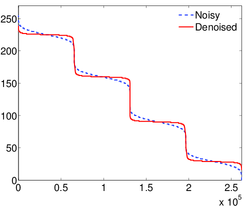}} 
 {\includegraphics[width=4cm,height=4cm]{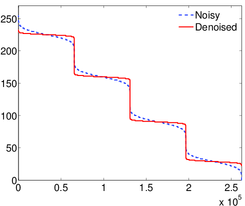}}  \\
  {\includegraphics[width=4cm,height=4cm]{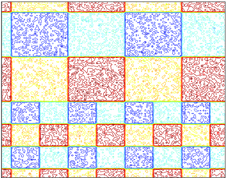}}
 {\includegraphics[width=4cm,height=4cm]{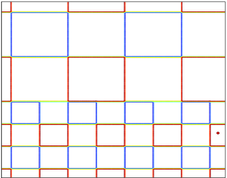}} 
  {\includegraphics[width=4cm,height=4cm]{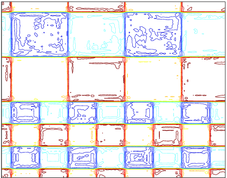}}
 {\includegraphics[width=4cm,height=4cm]{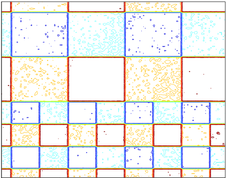}} \\
  \caption{{ Denoising experiment. Columns: noisy image, Neighborhood, NLM, and Bilateral filters. 
  Rows: noisy image, detail of the image, histograms, decreasing rearrangements, and 
  level curves of image details shown in row 2.}} 
\label{fig4}
\end{figure*}

In Figs.~\ref{fig2} to \ref{fig4} we show the results of applying these filters to the Boat, the Texture and the Squares images. The columns correspond to: noisy image, Neighborhood filter, Nonlocal means filter, and Bilateral filter. The rows correspond to: image, detail of the image, intensity histograms of noisy and denoised images, decreasing rearrangements of noisy and denoised images, level curves of image details showed in row 2. 

Although the Bilateral and the Nonlocal Means filters are applied only once, their execution time is always much larger than that of the iterated Neighborhood filter, 
for which we used the stopping criterium
\[
 \frac{\abs{J_*(v_{n+1}) - J_*(v_{n})}}{\abs{J_*(v_{n})}}<10^{-5},
\]
producing between eight iterations, for the Squares image, and twenty iterations, for 
the Texture image. We used the same parameter values for $h$ and $\rho$ in all the experiments.

As expected, the best visual result for the natural image is obtained with the Nonlocal Means filter: smoother and with a lower staircaising effect than the others. It is interesting to notice how the absence of local information in the Neighborhood filter 
produces regions with rapid intensity value changes, for instance in the clouds of 
the image. The smoothing effect of the local terms in the Bilateral and the NLM filters prevent the formation of this artifact.

A partial explanation of the worse behavior of the NF may be found in the corresponding plots for the histograms and the decreasing rearrangements. While the Bilateral and the NLM filters keep almost unchanged the gray intensity structure of the pixel mass, the NF concentrates most of the mass in few and disconnected values which, in general, is an undesired effect in natural images.

Finally, observe that all the filters produce a level lines shortening, notably the NLM filter.

For the Texture image, similar conclusions may be deduced. In this case, the level lines shortening is specially intense for the NF. The area between the circles is 
\emph{cleaned} to one single intensity value, around 225. We may check in the corresponding histograms the large difference between the mass assigned to this value
in the different filters. This is a first clue in the consideration of the NF as a
segmentation-like filter.

For the synthetic image Squares, the result of applying the NF is almost a perfect image recovery, while the Bilateral and the NLM filters keep always some noise due to 
the local diffusion. The spatial smoothening effect of the latter work against
denoising, for this image.

Let us finally point out to the border effects mentioned after Theorem~\ref{th.pde}, involving formula \fer{app.NFstar}, and related to the contrast loss induced by the NF. They 
are clearly visualized in the plots of the decreasing rearrangement of these images.
Also the anti-diffusive behavior of the algorithm, captured by the second order term of formula 
\fer{app.NFstar} is observed: concave 
regions induce increase on the iterate while convex regions induce decrease. The result is a steeper slope around the inflexion points at each iteration.

\subsection{ Numerical examples for segmentation}

 The intensity histogram, $h_u$, of an image $u$ is defined by the measure of its level sets
 \begin{equation*}
  h_u(q)=\abs{\{\by \in \O : u(\by)=q \}}, \qtext{for }q\in [\min u, \max u].
 \end{equation*}
 We therefore have the following relationship between the histogram and the distribution function of $u$,
 \begin{equation*}
  m_u(q)=\int_q^{\max u} h(s)ds.
 \end{equation*}
In particular, under regularity assumptions, critical points of the histogram coincides with inflexion points of the distribution function and, hence, of the decreasing rearrangement.

Observe that histogram critical points detection is the base for some segmentation algorithms, see for instance 
\cite{Tobias2002,Chang2006,Nath2011,Qin2011}. Due to the discrete nature of 
computations, finding the maxima of the histogram from where initiating 
a segmentation procedure is a challenging task.

At this respect, the NF may be seen 
as a way of detecting histogram maxima, i.e. 
inflexion points of the decreasing rearrangement, which 
produces an \emph{automatic} segmentation with the only tunning 
of the the window size controlled by $h$.

To demonstrate this capability, we applied the NF to MRI brain segmentation.
We used a phantom brain from the Simulated Brain Database  \cite{brainweb} 
with a $9\%$ of additive Riccian noise. 

\begin{figure*}[ht]
\centering
 {\includegraphics[width=4cm,height=4cm]{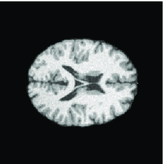}}
 {\includegraphics[width=4cm,height=4cm]{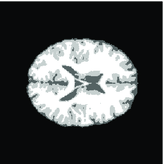}}
 {\includegraphics[width=4cm,height=4cm]{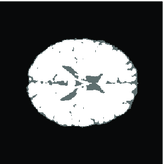}}
  {\includegraphics[width=4cm,height=4cm]{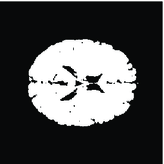}}\\
 {\includegraphics[width=4cm,height=4cm]{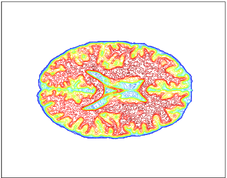}}
 {\includegraphics[width=4cm,height=4cm]{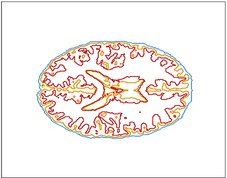}}
 {\includegraphics[width=4cm,height=4cm]{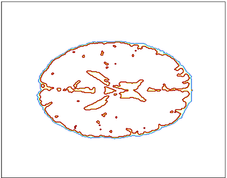}}
  {\includegraphics[width=4cm,height=4cm]{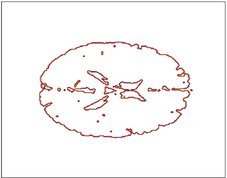}}\\
 \hspace{4cm}
 {\includegraphics[width=4cm,height=4cm]{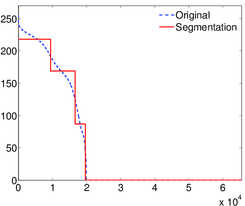}}
 {\includegraphics[width=4cm,height=4cm]{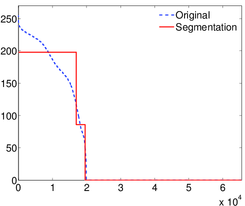}}
 {\includegraphics[width=4cm,height=4cm]{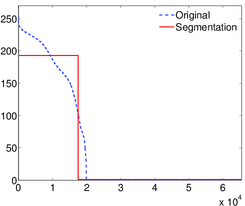}}\\
   \caption{{Segmentation experiment. Results of applying the Neighborhood filter with 
   several values of the window size $h$. Rows: Image, level curves and decreasing rearrangement, showing the number of segmented regions (flat regions). Columns: Image, 
   results of applying the NF with $h=17$, $h=20$, and $h=50$, respectively.}} 
\label{fig5}
\end{figure*}

In Fig.~\ref{fig5} we show an axial slice of the volume (initial image) an the 
corresponding segmentation in four, three and two regions reached by setting $h=17,~20,~50$, respectively. The contour lines and the decreasing rearrangement
are shown too.

\begin{figure*}[ht]
\centering
 \fbox{{\includegraphics[width=3.9cm,height=4cm]{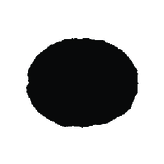}}}
 \fbox{{\includegraphics[width=3.9cm,height=4cm]{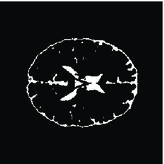}}}
 \fbox{{\includegraphics[width=3.9cm,height=4cm]{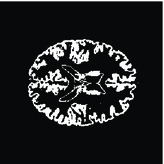}}}
  \fbox{{\includegraphics[width=3.9cm,height=4cm]{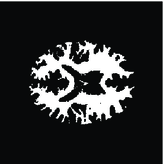}}}\\
  \hspace*{-4.35cm}\fbox{{\includegraphics[width=3.9cm,height=4cm]{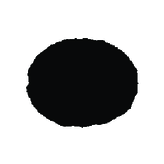}}}
 \fbox{{\includegraphics[width=3.9cm,height=4cm]{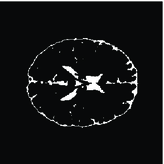}}}
 \fbox{{\includegraphics[width=3.9cm,height=4cm]{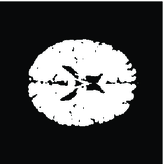}}}\\
   \caption{{Segmentation experiment. Masks of the segmented regions. First row: $h=17$. The NF produces four regions, corresponding to background, dura-mater and ventricles, grey matter and white matter. Second row: $h=20$. The NF produces three regions, corresponding to background, duramatter and ventricles, grey plus white matter.}} 
\label{fig6}
\end{figure*}

In Fig.~\ref{fig6} we show the masks of the segmented regions corresponding to $h=17,~20$.

\begin{figure*}[ht]
\centering
 {\includegraphics[width=5cm,height=4.5cm]{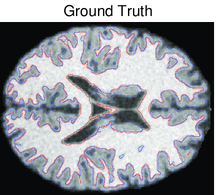}}
 {\includegraphics[width=5cm,height=4.5cm]{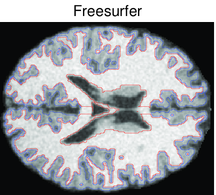}}
 {\includegraphics[width=5cm,height=4.5cm]{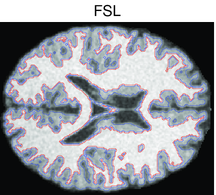}}\\
 {\includegraphics[width=5cm,height=4.5cm]{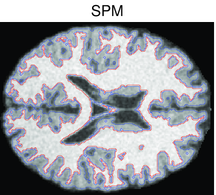}}
 {\includegraphics[width=5cm,height=4.5cm]{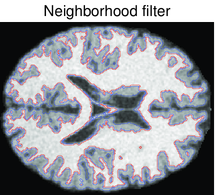}}
   \caption{{Segmentation experiment. Details of the segmentation produced with several 
   MRI standard packages and with the Neighborhood filter. See Table~\ref{tab:label} for details. }} 
\label{fig7}
\end{figure*}

In Fig.~\ref{fig7} we show the grey-white matter segmentation performed with the NF and with other  
standard packages: Freesurfer \cite{freesurfer}, FSL \cite{fsl} and SPM8 \cite{spm8}. 
The Dice coincidence coefficient is computed for all the algorithms, see Table~\ref{tab:label},  showing a good 
performance of the NF in relation to the more sophisticated algorithms implemented in 
the mentioned packages. The Dice coefficient is one if a perfect match to the ground truth
is attained. Zero, on the contrary. 

Although in Fig.~\ref{fig7} we have shown the results for one slice, the NF is applied directly to the whole volume, meaning that the dimension reduction is from a three 
dimensional space (the space of voxels) to a one dimensional space (the space of level lines measures). Thus, the time execution of the NF is several orders of magnitude lower than the others (a standard volume takes few seconds in a standard laptop). 

However, this is no more than a toy example, from where general conclusions can not be  inferred.

\begin{table}[!ht]
\centering
\caption{
\bf{Comparison among several algorithms}}
\begin{tabular}{|c||c|c|c|c||c|c|c|c|}
\hline\hline 
 & \multicolumn{4}{|c||}{Dice coefficient} 
 \\
 \hline  
             & Freesurfer      &   FSL  &   SPM  &    NF    \\\hline
 white       &  0.9490 & 0.9435 & 0.9468 & 0.9563 \\
 grey        &  0.8509 & 0.8599 & 0.8835 & 0.8797 \\
 \hline\hline  
\end{tabular}
\label{tab:label}
 \end{table}

\newgeometry{textwidth=18cm}

\section{Summary}
In this paper we introduced the use of the decreasing rearrangement 
to express nonlinear and nonlocal filters in terms of integral operators
in the one-dimensional space $[0,\abs{\O}]$.

We have proved properties related to the Neighborhood filter nonlinear iterative scheme. In particular, geometric properties like the invariance of level sets and the performance of the filter as a contrast change.

We have also proven a detailed qualitative behavior of the iterations as 
a power expansion in terms of the window size, and with coefficients 
which depend on up to second order derivatives of the iterations. This allowed us 
to distinguish two kind of effects of the filtering process: an anti-diffusive effect of shock-filter type, and a  contrast loss effect.

Motivated by the possible piece-wise constant steady state of the discrete problem, 
we have illustrated the interpretation of the filter as a segmentation algorithm, indeed
connected to other techniques involving the histogram thresholding.

The main conclusion of our work is that, for certain kind of images, among which those 
having concentrated their pixel mass around few intensity levels, the NF is appropriate both 
as a denoising and as a histogram-maxima based segmentation algorithm. The execution time 
of its rearranged version clearly out-performs those of other algorithms. However, for 
other kind of images, specially those with a relatively flat histogram, the results of the  NF are poor.

\section{Acknowledgments}
The authors thank to the anonymous reviewers for their interesting comments and suggestions, which have notably contributed to the improvement of our work.

The authors are partially supported by the Spanish DGI Project MTM2010-18427.

\end{document}